\pdfoutput=1
\documentclass[11pt]{article}

\usepackage{EMNLP2023}

\usepackage{times}
\usepackage{latexsym}
\usepackage{float}

\usepackage{booktabs, multirow} % for borders and merged ranges
\usepackage{soul}% for underlines
\usepackage[T1]{fontenc}
\usepackage[utf8]{inputenc}

\usepackage{microtype}

\usepackage{inconsolata}

\usepackage{todonotes}

\title{Too Much Information: Keeping Training Simple for BabyLMs}

\author{Lukas Edman \qquad  Lisa Bylinina
\vspace{.2cm}
 \\ Center for Language and Cognition \\ 
 University of Groningen \vspace{.1cm}
 \\ {\tt j.l.edman@rug.nl, e.g.bylinina@rug.nl}
}

\begin{document}
\maketitle
\begin{abstract}
This paper details the work of the University of Groningen for the BabyLM Challenge \cite{warstadt-et-al-2023-babylm}.
We follow the idea that, like babies, language models should be introduced to simpler concepts first and build off of that knowledge to understand more complex concepts. 
We examine this strategy of simple-then-complex through a variety of lenses, namely context size, vocabulary, and overall linguistic complexity of the data. 
We find that only one, context size, is truly beneficial to training a language model.
However this simple change to context size gives us improvements of 2 points on average on (Super)GLUE tasks, 1 point on MSGS tasks, and 12\% on average on BLiMP tasks. 
Our context-limited model outperforms the baseline that was trained on 10$\times$ the amount of data.

% We find that a combination of vocabulary expansion from the character level up to the subword level as well as the incremental increase of context size improves the performance of language models on the BLiMP benchmark by about 9\% over the provided baseline. We also find that such methods can be used with larger models of up to 600M parameters \todo{check the parameter count} (Deberta-Large) and still see further improvements up to 12\% over the baseline, showing that one can still achieve convergence with a large model trained from scratch even in low-resource settings.
\end{abstract}

\section{Introduction}

The pretraining of language models has traditionally relied on large amounts of data, which, for many languages, is readily available. However there exist several low-resource languages in which even unlabeled data is not so readily available. While transferring knowledge from other languages is often an effective way to achieve better performance, there may be implicit biases also transferred from the text of the higher-resource language, which could be potentially harmful. Additionally, given that a 13 year old sees less than 100 million words in their lifetime (orders of magnitude less than the amount used in LM pretraining), there ought to be methods that more efficiently learn from limited data.

Such is the motivation for the BabyLM Challenge and subsequently our work. We focus on the \texttt{strict-small} track, which limits the training data to only 10 million words, from a selection of domains with varying complexity (from child speak up to Wikipedia articles). 

In our work, we investigate different methods for introducing the model to varying levels of complexity. Namely we ramp up the difficulty of the pretraining along 3 avenues:
\begin{enumerate}
    \item Context length
    \item Dataset complexity
    \item Vocabulary size
\end{enumerate}

Concerning context length, we adopt the strategy of starting with a small number of tokens per input and increasing this over the course of training, with the intuition that a human typically learns a language starting with short sentences with limited cross-sentential context, and builds up from there to longer contexts.

In addition, the sentences initially learned by a human are also simpler conceptually, starting with frequently-used words and building up to rarer words. To this end, we develop a strategy to filter the dataset such that the model starts training on simpler data and later trains on more complex data.

Similarly, we also follow the intuition that a human develops a vocabulary over time, originating from the chunking of characters within the words, and as such we start with a character-level vocabulary and introduce a transfer method to give a good initialization for a larger subword vocabulary.

\section{Related Work}
Concerning context size, prior work \cite{edman2022importance} has shown that in low-resource language modeling, using a lower context size can greatly help with model convergence. The concept of increasing context size is not novel: BERT \cite{devlin2018bert} was initially trained on a smaller context size of 128 tokens before being increased to 512, though, to our knowledge, this was done for efficiency reasons. There have been several works on internally reducing the scope of contextualization by limiting attention to local patches \cite{beltagy2020longformer,zaheer2020big}, thereby decreasing the complexity of self-attention. These works were done with processing long documents in mind, however, and can have a negative impact on model speed given an extra layer of complexity in calculating self-attention. 

Concerning vocabulary size, there is ample work on character-level models, where they have been shown to require less data for pretraining while achieving the same or better performance at the cost of training and inference speed \cite{xue2022byt5}. Character models also can greatly outperform subword models on out-of-domain tasks \cite{boukkouri2020characterbert}, low-resource translation \cite{edman2023character}, and tasks which require morphology or character-level perturbations \cite{xue2022byt5, ingolfsdottir2023byte}. Their performance in these scenarios has been largely attributed to their non-static vocabulary, allowing for good initializations to unseen or rarely-seen words. 
All of this points to character-informed models being potentially useful for this shared task. 

Concerning lexical complexity, \cite{eldan2023tinystories} has shown that using a synthetic dataset of children's stories, written for a 3 or 4 year old to understand, one can train a small (<10M parameter) Transformer model and generate stories near the quality of much larger models. 

Another group of NLP approaches that condition learning on linguistic complexity is a branch of curriculum learning, exploring potential benefits from exposing models to training samples in a meaningful order, from easy to hard (\citealt{bengio2009curriculum, kocmi2017curriculum, zhang2018empirical} among many others). These approaches show conceptual promise but are complicated by the choice of appropriate complexity measures and the pacing function.\footnote{Pacing function is a broad term used by \citet{soviany2022curriculum}, describing the method for ramping up difficulty across the course of training.}

\section{Method}
\subsection{Model Choice}
We opted to use encoder-only models. We initially experimented with encoder-decoder models, but found that the evaluation metrics for this shared task being non-generative gave encoder-only models an advantage, as it allows for full attention, rather than only causal attention.
In terms of specific model selection, we opted for RoBERTa-base \cite{liu2019roberta} in order to directly compare with the provided baseline. We also experimented with (and ultimately submitted) DeBERTa-large \cite{he2021debertav3} as it is a larger model and considered state-of-the-art for encoder-only models. 

\subsection{Training and Evaluation}
Our pretraining uses the standard MLM scheme \cite{liu2019roberta}, which proved most effective initial tests.\footnote{We also varied masking amounts to 20\% and 40\% following \citet{wettig2022should}, but did not see any increased performance on BLiMP.} Table \ref{tab:hyperparams} shows the hyperparameters we used for our pretraining experiments. For fine-tuning, we use the default hyperparameters provided by the shared task organizers.

\begin{table}[!htp]\centering
\begin{tabular}{lrr}\toprule
Hyperparameter &Value \\\midrule
Learning rate &1e-4 \\
Decay &0.01 \\
Warmup steps & 10000 \\
Optimizer &AdamW \\
Batch size &256 \\
Epochs &50 \\
\bottomrule
\end{tabular}
\caption{Hyperparameters used.}\label{tab:hyperparams}
\end{table}

We primarily evaluate with BLiMP \citep{warstadt2020blimp}, due to its speed of evaluation and not requiring a fine-tuning step. We also report results of our best models for the BLiMP supplement, (Super-)GLUE \citep{wang2018glue,wang2019superglue}, and MSGS \citep{warstadt-etal-2020-learning} tasks.

\subsection{Vocabulary size}
We first experiment with vocabulary size. For creating the vocabulary, we use SentencePiece's Unigram model \cite{kudo2018sentencepiece,kudo2018subword}. We found that a vocabulary size of 40k provided the best standalone performance on BLiMP (we report this in Appendix \ref{sec:appendix}).

We further experiment with a character-level vocabulary, and transferring to a subword vocabulary (of size 40k). To enable this transfer, we copy over all character-only embeddings, and initialize subword embeddings as the sum of their respective character embeddings. The main body of the transformer model is also directly copied. 
The language modelling head is simply re-trained from scratch.

\subsection{Context size}
We also experiment with context sizes in powers of 2, from 16 to 256. To achieve a consistent and coherent context size, we split the data into $n$-token examples (with $n$ being the context size), prior to shuffling.  
Our initial experiments with determining the optimal vocabulary size use a context size of 64, although we later find that a context size of 32 performs slightly better.

\subsection{Curriculum learning}

We explored potential gains from different order of exposure of the model to training data, inspired by curriculum learning approaches (see \citealt{bengio2009curriculum} and much subsequent work; for a recent comprehensive survey of the field of curriculum learning, see \citealt{soviany2022curriculum}). 

The basic motivating intuition is to start the training with subsets of data that are `simpler' than others in some relevant sense, gradually increasing the complexity of data the model is trained on. Hopefully, simple data can give the model a head start that would also form a foundation for linguistic generalization. To try out this idea, we formulate a {\bf complexity measure} that we use in data reordering. The measure is a combination of the following features:

\begin{itemize}
    \item {\bf Type/Token Ratio}: The number of unique words in a text divided by the length of the text in words. The feature targets lexical diversity of the text per text unit.
    \item {\bf Mean word rarity}: The mean of rarity scores across all words in the text (word rarity score is 1 - normalized log-frequency; it ranges from 0 to 1, the higher the rarer). This is another measure of text complexity via lexical diversity -- this time, based on how rare the words used in the text are, as judged based on the whole training dataset.  
    \item {\bf Max word rarity}: The maximum of word rarity scores in the text. Same as above, but picking out the maximum -- the peak of complexity-as-rarity reached in the text.
    \item {\bf Punctuation density}: The proportion of punctuation marks in the union of words and punctuation marks in the text. This proportion is used as a proxy to syntactic complexity. 
    \item {\bf Mean sentence length} in the text, in words.
    \item {\bf Mean word length} in the text, in characters. These last two scores approximate syntactic and morphological/lexical complexity, respectively.
\end{itemize}

Features like these and their different combinations are often used to measure text complexity and/or readability
\citep{bengio2009curriculum, spitkovsky2009baby, cirik2016visualizing,kocmi2017curriculum,zhang2018empirical,platanios2019competence,chang2021does}.

In our experiments, we scale all these features to fit into the [0,1] interval (with MinMax scaler) and use their mean as our complexity measure.

To assess the role of data ordering along the complexity scale based on the measure above, we trained triples of minimally different models, keeping everything apart from the data ordering fixed: 
\begin{itemize}
    \item {\bf Curriculum model}: All training data is ordered by increasing complexity. 
    \item {\bf No-curriculum model}: No particular order is imposed on the training data.
    \item {\bf Reversed-curriculum model}: Training data is ordered by {\bf decreasing} complexity.
\end{itemize}

All models in this set of experiments are RoBERTa-base models trained following the two-stage procedure described in Section 4.1 -- first, the models are trained on context size 32, then the context is increased to 128. Unlike in other experiments, however, each of the stages was further divided into three consecutive phases:
\begin{itemize}
    \item {\bf Phase 1}: The first 1/3 of the data is used in training, the other 2/3 are withheld. The curriculum model just sees the `easiest' data here; the reversed-curriculum model sees the `most difficult' portion; the baseline, no-curriculum model sees 1/3 of data without any particular selection;
    \item {\bf Phase 2}: Another 1/3 of the data is unlocked. Now all models are being trained on 2/3 of all training data. Both the curriculum model and the reversed-curriculum model now have access to the middle of the complexity range. 
    \item {\bf Phase 3}: The final 1/3 of data is unlocked. Now all models are being trained on the whole range of complexity.
\end{itemize}

The data-unlocking procedure above happens twice: first, on a small context size (32 tokens), and later when the context size is increased (128 tokens). 

Using the taxonomy of curriculum learning in \cite{soviany2022curriculum}, we can describe our approach as vanilla data-level curriculum learning with easy-then-hard iterative schedule.

% To combine both the vocabulary expansion and context size increase, we found the most intuitive way to do this was so start by training a character-level model with a context of 64 character-level tokens, transferring to a context of 64 subword-level tokens and training again, and then finally training a 3rd time to a context of 256 subword-level tokens. 

\section{Results}

% \begin{table}[!htp]\centering
% \scriptsize
% \begin{tabular}{lrrrrrr}\toprule
% &\multicolumn{5}{c}{Context size} \\\cmidrule{2-6}
% BLiMP Scores (\%) &16 &32 &64 &128 &256 \\\midrule
% Anaphor agreement &92.4 &\textbf{94.6} &91.8 &84.0 &85.4 \\
% Argument structure &72.9 &74.4 &\textbf{74.6} &66.2 &61.5 \\
% Binding &70.8 &70.8 &\textbf{71.5} &64.4 &65.6 \\
% Control raising &71.8 &\textbf{72.0} &70.9 &63.3 &58.2 \\
% Determiner noun agreement &95.8 &\textbf{96.4} &95.4 &88.8 &56.4 \\
% Ellipsis &62.8 &82.9 &\textbf{88.6} &72.6 &64.5 \\
% Filler gap &\textbf{75.6} &75.0 &72.0 &62.7 &69.3 \\
% Irregular forms &86.2 &92.4 &\textbf{92.6} &85.8 &60.7 \\
% Island effects &\textbf{60.8} &58.9 &50.9 &38.5 &50.9 \\
% NPI licensing &70.2 &71.8 &\textbf{73.0} &46.7 &41.3 \\
% Quantifiers &\textbf{72.2} &70.4 &70.9 &58.1 &52.7 \\
% Subject verb agreement &83.8 &\textbf{84.1} &81.3 &61.8 &53.1 \\\midrule
% Average &76.3 &\textbf{78.6} &77.8 &66.1 &60.0 \\
% \bottomrule
% \end{tabular}
% \caption{Performance of varying context sizes on Roberta-base. Best in bold.}\label{tab:ctx_size}

% \end{table}

\begin{figure}
    \centering
    \includegraphics[scale=0.5]{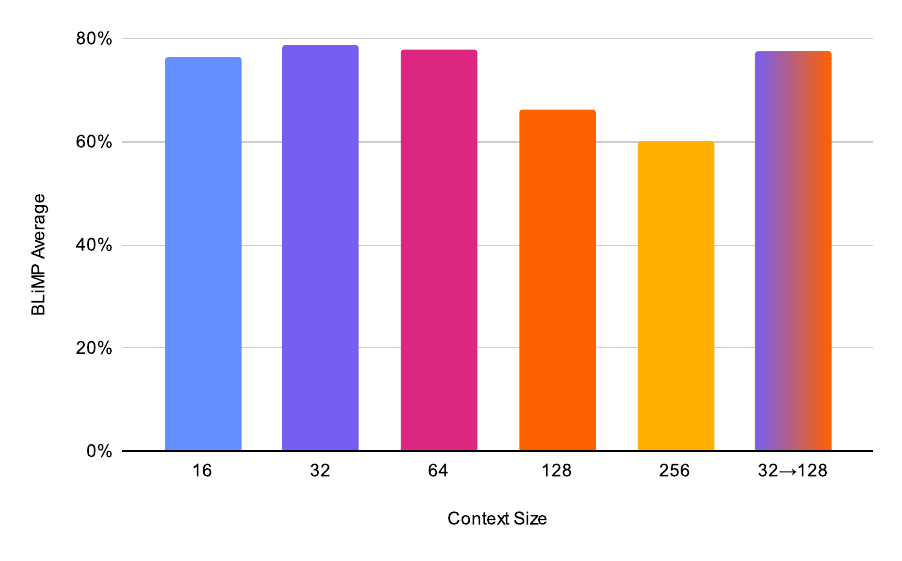}
    \caption{Average BLiMP score for models trained using various context sizes. 32$\rightarrow$128 indicates a model trained initially on context size 32, then trained again on 128. }
    \label{fig:ctx_size}
\end{figure}

\begin{figure}
    \centering
    \includegraphics[scale=0.5]{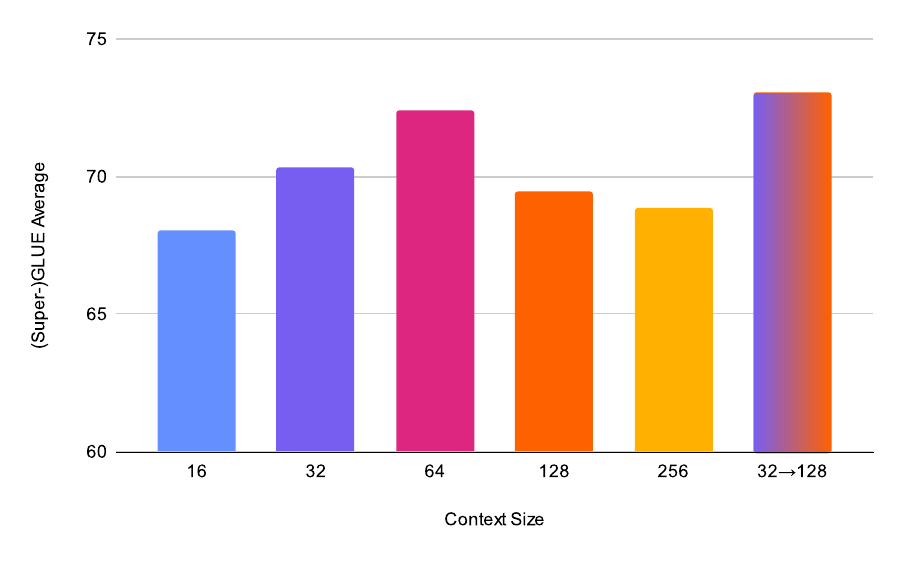}
    \caption{Average (Super-)GLUE score for models trained using various context sizes. 32$\rightarrow$128 indicates a model trained initially on context size 32, then trained again on 128. }
    \label{fig:ctx_size_glue}
\end{figure}

\subsection{Context Size}
The vast majority of our improvement comes from limiting the context size. We show this in Figures \ref{fig:ctx_size} and \ref{fig:ctx_size_glue}. 
Here we can see that a context size of 32 gives the best performance on BLIMP, whereas 64 gives the best performance on GLUE. The overall shift in trend between the two benchmarks fits with the fact that the average input length is longer in GLUE than in BLIMP. There is a substantial drop in performance using a context size of greater than 64. 
To our understanding, the baselines provided by the task organizers use a context size of 128, which may explain their relatively poorer performance (as shown later in Figure \ref{fig:final}). 

However, if we simply first train with a context size of 32, then increase the context size to 128, we see a substantial gain over training on 128 from the beginning. In the case of GLUE, we see that increasing the context size from 32 to 128 increases the performance beyond what simply training on 32 \textit{or} 128 alone can accomplish. This suggests that a larger context size is indeed necessary for performance on (Super-)GLUE, but pretraining initially on a smaller context can guide the model to more efficient training on larger context sizes. 

% Nevertheless, even the baseline trained on 100M words performs worse than our models with smaller context size trained only on 10M words. This shows just how important limiting context size can be when pretraining a low-resource language model. 
\subsection{Vocabulary Expansion}
Next, we look at the performance of our models which were initially trained on a character-level vocabulary, then transferred to our 40k subword vocabulary. We show the results in Table \ref{tab:vocab}.

\begin{table}[!htp]\centering
\begin{tabular}{lrrrr}\toprule
& &\multicolumn{2}{c}{Vocabulary size} \\\cmidrule{3-4}
& &40k &Char$\rightarrow$40k \\\midrule
\multirow{2}{*}{Context size} &32 &\textbf{78.6} &77.1 \\
&64 &77.8 &\textbf{78.6} \\
\bottomrule
\end{tabular}
\caption{Average performance on BLiMP across context and vocabulary sizes.}\label{tab:vocab}
\end{table}

As we can see, the performance is mixed and depends on the context size. For context size 64, there appears to be an improvement, however for context size 32, the performance drops. The lack of improvement for context size 32 led us to leave out this technique in our final model, as the potential gains are inconsistent and training first on the character level adds a costly extra pretraining step.

As for the use of characters in low-resource pretraining, we suspect that there are better ways of integrating rather than via an extra initial pretraining step. Using our method, the model is susceptible to forgetting what it has learned during the character-level pretraining when it is pretraining for the second time.

Additionally, the evaluation metrics chosen for this shared task do not stand out as tasks where character models would be particularly beneficial. Other tasks where character-level models have been shown to greatly outperform subword-level models such as morphological inflection would be perhaps more suitable for assessing the potential benefits of our character-informed model.

% , and we see a small but noticeable improvement. As expected, our performance on quantifiers improves, but the model also improves overall on most other subtasks. To test whether a smoother transition from characters to subwords would improve performance, we also tried adapting first from characters to a subword vocabulary of 8k, and then repeating the process from 8k to 40k. This led to worse performance, possibly because more intermediate steps only increases the chances of catastrophic forgetting. 

\subsection{Curriculum learning}

We evaluate the results of data ordering from simple to complex against two alternatives: no data ordering and reversed data ordering (from difficult to simple). We train triples of models that minimally differ from each other -- everything apart from the order of data is kept constant. 

\begin{figure}
    \centering
    \includegraphics[width=.48\textwidth]{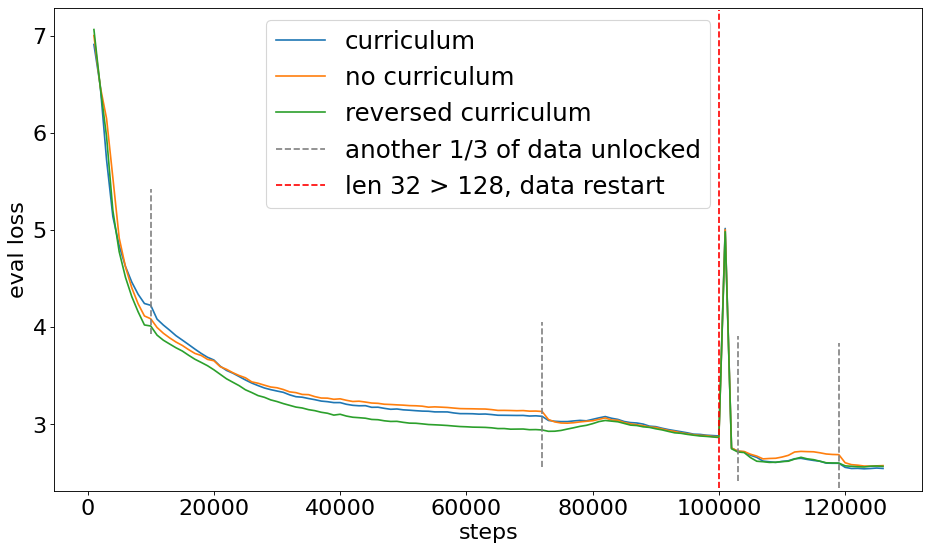}
    \caption{Loss dynamics for three minimally different models: curriculum; no curriculum; reversed curriculum.}
    \label{fig:curriculum}
\end{figure}

Figure \ref{fig:curriculum} shows evaluation loss dynamics of one typical model triples during training (we set up several training experiments, varying the number of epochs per phase, without qualitative change in results, so here we only report one of them).

While there are stages in training where the loss seems to indicate an advantage of the curriculum model against the baseline one, the no-curriculum model eventually catches up. Perhaps more surprisingly, the reversed-curriculum model shows systematically lower loss during longer phases of training.

Targeted linguistic evaluation also shows mixed results. Table \ref{tab:curric} lists BLiMP scores for the model triple:

%\begin{adjustwidth}{-2.5 cm}{-2.5 cm}\centering\begin{threeparttable}[!htb]...\end{threeparttable}\end{adjustwidth}

\begin{table}[!htp]\centering
\begin{tabular}{lrrrr}\toprule
BLiMP Scores &Cur. &Rev. cur. &No cur. \\\midrule
Anaph. agree. &89.4 &89.4 &\textbf{90.1} \\
Arg. struct. &\textbf{73.6} &71.3 &72.5 \\
Binding &69.3 &\textbf{70.7} &68.8 \\
Control raising &\textbf{71.1} &71.0 &70.5 \\
Det. noun agree. &\textbf{95.3} &94.0 &\textbf{95.3} \\
Ellipsis &\textbf{86.2} &82.2 &83.7 \\
Filler gap &\textbf{75.9} &70.7 &72.8 \\
Irregular forms &83.7 &83.0 &\textbf{86.5} \\
Island effects &54.9 &\textbf{62.1} &53.5 \\
NPI licensing &67.1 &66.8 &\textbf{70.5} \\
Quantifiers &68.9 &\textbf{69.3} &66.2 \\
SV agree. &81.2 &\textbf{81.8} &78.8 \\ \midrule
Average &\textbf{76.4} &76.0 &75.8 \\
\bottomrule
\end{tabular}
\caption{The effect of data ordering on linguistic generalization.}\label{tab:curric}
\end{table}

We don't see a clear pattern in what types of linguistic phenomena benefit from a particular order of data exposure and can't conclude whether the observed effects are robust and systematic.  

The lack of a clear benefit from the curriculum might be traced back to at least one of the following:

\begin{itemize}
    \item Low quality of the complexity metric;
    \item Inadequacy of the complexity metric for the training objective;
    \item Interfering data noise; 
    \item Non-optimal pacing function;
    \item The genuine lack of advantage from data reordering.
\end{itemize}

To illustrate some of these considerations, we pick three typical samples from the ordered dataset for context size 32 -- from the `simplest' end, from the middle, and from the `most complex' end, respectively: 

\begin{enumerate}
    \item[(a)]    \tt{down!\\
 up up up up up up up up up \\up up up up up down!}
\item[(b)] \tt{the flared skirt of the cone yet to be combed, and this provide}
\item[(c)] \tt{p;amp;gt;\&amp;amp;gt;Exactly. \&amp;amp;gt;\&amp;amp;gt;Combining}
\end{enumerate}

The easiest samples are indeed linguistically simple -- they contain a lot of repetitions, very simple syntactic structures and very frequent words. At the same time they are not very representative of the rest of the dataset, both grammatically and lexicaly. The typical sample from the main body of the dataset -- samples like (b) -- do not show the characteristic repetitive pattern and a large proportion of the lexical material across the dataset falls outside of what the simplest samples contain. The simplest data defined the way we do it is useful for generalization to the rest of the data only to a very limited degree: the model does see the most frequent words, but the contexts of their use are pretty different from how they are typically used elsewhere. For a model with non-character-level tokenization, it might not be particularly helpful.

On the other side of the complexity scale, a lot of samples are indeed difficult, but in a way that does not necessarily reflect true linguistic complexity: vocabulary and punctuation features push up samples that contain elements of HTML, have collapsed space symbols, are lists or are written in languages that are not the main language of the dataset. 

In a sense, both extreme ends of the complexity scale contain samples that are probably not good grounds for linguistic generalization given the MLM training objective, but in different ways. 

\subsection{Model Size}
Table \ref{tab:model_size} shows the performances of the two models we used, as well as DeBERTa-base to control for the differences in model architecture between RoBERTa and DeBERTa. We can see that DeBERTa-large generally performs best. Interestingly, we see that switching from RoBERTa to DeBERTa seems to account for the difference in GLUE scores, but scaling up to large accounts for the increase in BLiMP scores. This shows that when limiting the context size, we can potentially scale up to larger models even when data is scarce.

\begin{table}[!htp]\centering
\begin{tabular}{lrrr}\toprule
&Ro-base & De-base &De-large \\\midrule
BLiMP &78.6 & 79.0 &\textbf{81.0} \\
BLiMP supp. &\textbf{63.8} & 59.8 &\textbf{63.8} \\
MSGS & -70.7 & -62.2 & \textbf{-53.7} \\
GLUE &70.3 & \textbf{72.5}&\textbf{72.5} \\
\bottomrule
\end{tabular}
\caption{RoBERTa-base versus DeBERTA-base and large on all tasks. MSGS is the average Matthew's Correlation Coefficient multiplied by 100. Best in bold.}\label{tab:model_size}
\end{table}

We also experimented with training a Deberta-XL model, which is identical to Deberta-Large except with 48 layers rather than 24. Our results on BLiMP were however not better (roughly 2\% worse than the comparable \texttt{large} model), so it would seem that there is a limit to how much one can simply scale up model size and see performance improvements when it comes to pretraining on limited data.

\subsection{Submission}

\begin{figure}
    \centering
    \includegraphics[scale=0.37]{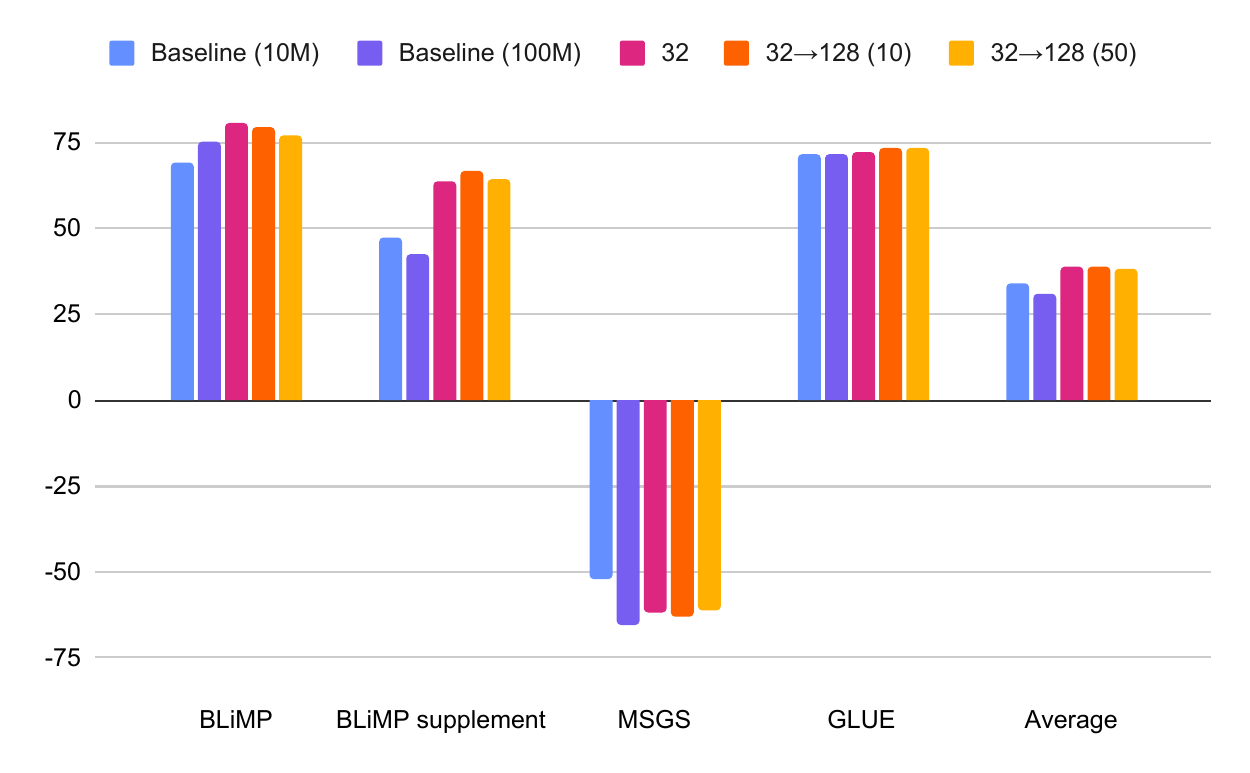}
    \caption{Average scores for submitted models compared to baselines. 32$\rightarrow$128 indicates a model trained initially on context size 32, then trained again on 128. The number in parentheses indicates the number of epochs trained on for the second iteration of pretraining. MSGS scores are the average Matthew's Correlation Coefficient, multiplied by 100. }
    \label{fig:final}
\end{figure}

In Figure \ref{fig:final}, we show the overall results for our best models, compared to the baselines. We also report results on each individual sub-task in Appendix \ref{sect:all_final}. Our final models include a model trained only on context size 32, and two trained again on context size 128, one for 10 epochs and one for 50 epochs. As our one trained with 10 additional epochs performed best on average, this was our final submission.
We can see the trade-off for context size between the GLUE and BLiMP scores, as BLiMP favors models trained on a shorter context while GLUE favors models trained on a longer context. MSGS appears to also have some slight preference for models trained on a shorter context, though the differences between all models is comparatively small. Interestingly, the 10M baseline is better on average than the 100M baseline on MSGS, as well as the BLiMP supplement.
We see the largest difference in the BLiMP supplement, where our models outperform the baselines by around 20 points on average. Much of this improvement comes from the \texttt{qa\_congruence\_easy} set, where our best model achieved a score of 81\%, compared to the baseline score of 31\%. 

\section{Conclusion}
Our conclusion is very simple: if you want to pretrain a model on little data, train with a smaller context size. This can greatly aid in model convergence such that no specific hyperparameter tuning or complex methods need to be used for superior performance.

In fact, both of our more ``complex'' approaches concerning initialization with a character vocabulary and curriculum learning proved to be unreliable, where gains paled in comparison to the gains realized from simply lowering context size. 

If a larger context size is eventually needed, such as for some GLUE tasks, continuing training with a larger context size can provide some benefit.
We do think that there may be a smarter way to control context size, such as a gradual increasing during training, which could lead to smoother and faster training. Additionally we expect that there are other potential ways to implicitly limit context size, such as limiting self-attention, which may achieve a similar effect.

%Please add the following packages if necessary:
%If the table is too wide, replace \begin{table}[!htp]...\end{table} with
%\begin{adjustwidth}{-2.5 cm}{-2.5 cm}\centering\begin{threeparttable}[!htb]...\end{threeparttable}\end{adjustwidth}

\section*{Acknowledgements}
We thank the Center for Information Technology of the University of Groningen for their support and for providing access to the Hábrók high performance computing cluster.
\bibliography{anthology,custom}
\bibliographystyle{acl_natbib}

\appendix

\section{Vocabulary Size}
\label{sec:appendix}

\begin{table*}[!htp]\centering
\begin{tabular}{lrrrrrrrrr}\toprule
&\multicolumn{8}{c}{Vocabulary Size} \\\cmidrule{2-9}
BLiMP Scores (\%) &Char &8k &16k &24k &32k &40k &48k &64k \\\midrule
Anaphor agreement &44.0 &88.3 &90.1 &\textbf{92.9} &92.6 &91.8 &92.8 &91.3 \\
Argument structure &59.4 &69.0 &73.8 &73.6 &73.6 &74.6 &74.4 &\textbf{74.7} \\
Binding &61.5 &69.2 &69.3 &70.4 &69.3 &\textbf{71.5} &68.9 &68.3 \\
Control raising &60.0 &63.0 &68.2 &69.1 &69.6 &70.9 &\textbf{71.7} &69.5 \\
Determiner noun agreement &89.2 &89.5 &88.0 &89.8 &94.5 &95.4 &\textbf{96.6} &96.5 \\
Ellipsis &42.4 &85.8 &84.9 &87.1 &86.4 &\textbf{88.6} &84.5 &87.3 \\
Filler gap &70.3 &\textbf{73.9} &73.0 &73.7 &73.0 &72.0 &74.0 &73.5 \\
Irregular forms &78.9 &84.4 &89.6 &89.3 &89.6 &\textbf{92.6} &85.8 &88.8 \\
Island effects &43.9 &44.4 &46.8 &48.9 &51.8 &50.9 &53.0 &\textbf{53.4} \\
NPI licensing &55.0 &56.0 &63.5 &68.3 &70.2 &\textbf{73.0} &67.0 &67.1 \\
Quantifiers &\textbf{80.4} &66.5 &70.8 &68.3 &69.0 &70.9 &71.0 &68.5 \\
Subject verb agreement &71.4 &78.2 &79.3 &80.3 &\textbf{83.5} & 81.3 &81.7 &81.1 \\ \midrule
Average &63.0 &72.3 &74.8 &76.0 &76.9 &\textbf{77.8} &76.8 &76.7 \\
\bottomrule
\end{tabular}
\caption{BLiMP scores for each vocabulary size tested. ``Char'' refers to a character-level model. Best in bold.}\label{tab:vocab_size}

\end{table*}

We experiment with vocabulary size, as shown in Table \ref{tab:vocab_size}. Here, we initially chose a context size of 64, which we later show to be a close to optimal. The results favor a vocabulary size of 40k, however we note that certain aspects of the character model, namely its performance on quantifiers, indicates that it could complement the subword vocabulary. 

\section{Final Full Results} \label{sect:all_final}
We report the results for the baselines and our submitted models in Table \ref{tab:all_final}.

\begin{table*}[!htp]\centering
\scriptsize
\begin{tabular}{lrrrrrrr}\toprule
& &Baseline (10M) &Baseline (100M) &32 &32$\rightarrow$128 (10) &32$\rightarrow$128 (50) \\\midrule
\multirow{12}{*}{BLiMP} &Anaphor agreement &81.5 &89.5 &\textbf{94.5} &93.0 &88.0 \\
&Argument structure &67.1 &71.3 &\textbf{76.3} &74.5 &72.9 \\
&Binding &67.3 &71.0 &\textbf{77.0} &76.3 &74.9 \\
&Control raising &67.9 &67.1 &\textbf{75.5} &74.2 &72.8 \\
&Determiner noun agreement &90.8 &93.1 &\textbf{95.6} &94.4 &91.0 \\
&Ellipsis &76.4 &83.8 &\textbf{84.1} &78.5 &77.4 \\
&Filler gap &63.5 &68.0 &\textbf{80.0} &78.8 &76.0 \\
&Irregular forms &87.4 &89.6 &\textbf{87.9} &85.8 &83.2 \\
&Island effects &39.9 &54.5 &68.4 &\textbf{70.7} &68.8 \\
&NPI licensing &55.9 &66.3 &72.5 &\textbf{73.2} &69.9 \\
&Quantifiers &70.5 &70.3 &\textbf{70.8} &66.4 &66.0 \\
&Subject verb agreement &65.4 &76.2 &\textbf{89.0} &87.8 &84.3 \\ \midrule
\multirow{5}{*}{BLiMP Supp.} &hypernym &49.4 &\textbf{50.8} &46.9 &49.1 &45.4 \\
&qa\_congruence\_easy &31.3 &34.4 &76.6 &\textbf{81.3} &73.4 \\
&qa\_congruence\_tricky &32.1 &34.5 &45.5 &\textbf{49.1} &46.7 \\
&subject\_aux\_inversion &71.7 &45.6 &82.8 &\textbf{84.3} &83.3 \\
&turn\_taking &53.2 &46.8 &67.1 &68.9 &\textbf{73.6} \\ \midrule
\multirow{11}{*}{GLUE} &CoLA &70.8 &75.9 &\textbf{76.8} &\textbf{76.8} &77.4 \\
&SST-2 &87.0 &\textbf{88.6} &87.8 &\textbf{88.6} &88.0 \\
&MRPC (F1) &79.2 &\textbf{80.5} &70.6 &72.9 &73.5 \\
&QQP (F1) &73.7 &78.5 &86.6 &86.6 &\textbf{87.1} \\
&MNLI &73.2 &68.7 &76.4 &76.2 &\textbf{77.1} \\
&MNLI-mm &74.0 &\textbf{78.0} &77.3 &76.3 &77.0 \\
&QNLI &77.0 &82.3 &83.2 &\textbf{83.5} &79.7 \\
&RTE &\textbf{61.6} &51.5 &50.5 &55.6 &56.6 \\
&BoolQ &66.3 &59.9 &65.2 &\textbf{67.9} &67.2 \\
&MultiRC &61.4 &61.3 &61.9 &62.0 &\textbf{64.4} \\
&WSC &61.4 &61.4 &\textbf{61.5} &\textbf{61.5} &\textbf{61.5} \\ \midrule
% \multirow{11}{*}{MSGS} &main\_verb\_control &84.1 &93.0 &\textbf{100.0} &\textbf{100.0} &\textbf{100.0} \\
% &control\_raising\_control &\textbf{100.0} &\textbf{100.0} &92.2 &96.9 &96.8 \\
% &syntactic\_category\_control &99.4 &\textbf{100.0} &85.5 &86.9 &88.5 \\
% &lexical\_content\_the\_control &93.5 &\textbf{100.0} &\textbf{100.0} &\textbf{100.0} &\textbf{100.0} \\
% &relative\_position\_control &96.4 &89.0 &99.3 &95.8 &\textbf{99.4} \\
% &main\_verb\_lexical\_content\_the &67.7 &\textbf{68.3} &66.6 &66.7 &66.6 \\
% &main\_verb\_relative\_token\_position &68.6 &66.8 &78.2 &\textbf{80.4} &72.3 \\
% &syntactic\_category\_lexical\_content\_the &66.7 &66.6 &\textbf{75.7} &75.4 &73.9 \\
% &syntactic\_category\_relative\_position &68.6 &\textbf{80.2} &65.6 &64.9 &64.7 \\
% &control\_raising\_lexical\_content\_the &\textbf{84.2} &67.4 &66.5 &67.2 &67.9 \\
% &control\_raising\_relative\_token\_position &65.7 &67.4 &\textbf{73.2} &68.9 &69.3 \\
\multirow{6}{*}{MSGS} &CR\_LC &\textbf{-0.28} &-0.89 &-0.98 &-0.92 &-0.49 \\
&CR\_RTP &-0.78 &-0.91 &\textbf{-0.52} &-0.85 &-0.84 \\
&MV\_LC &\textbf{-0.99} &-1.00 &-1.00 &-1.00 &-1.00 \\
&MV\_RTP &-0.79 &-0.15 &-0.32 &\textbf{-0.18} &-0.60 \\
&SC\_LC &\textbf{0.16} &-0.58 &-0.38 &-0.29 &-0.18 \\
&SC\_RP &-0.45 &\textbf{-0.39} &-0.51 &-0.53 &-0.55 \\ \midrule
\multirow{4}{*}{AoA} &Overall &2.06 &2.06 &2.06 &\textbf{2.05} &\textbf{2.05} \\
&Nouns &\textbf{1.99} &\textbf{1.99} &2.00 &\textbf{1.99} &2.00 \\
&Predicates &1.85 &\textbf{1.82} &1.85 &1.85 &1.83 \\
&Function words &2.65 &2.66 &2.60 &2.58 &\textbf{2.55} \\
\bottomrule
\end{tabular}
\caption{All individual results for our final models, versus the baselines. Best in bold. }\label{tab:all_final}
\end{table*}

\end{document}